\documentclass{article}

\usepackage{arxiv}

\usepackage[utf8]{inputenc} 
\usepackage[T1]{fontenc}    
\usepackage{hyperref}       
\usepackage{url}            
\usepackage{booktabs}       
\usepackage{amsfonts}       
\usepackage{nicefrac}       
\usepackage{microtype}      
\usepackage{lipsum}
\usepackage{graphicx}
\graphicspath{ {./images/} }

\title{HiLight: Technical Report on the Motern AI Video Language Model}

\author{
 Zhiting Wang \\
  \texttt{wangzt@motern.com} \\
   \And
 Qiangong Zhou \\
  \texttt{zhouqg@motern.com} \\
  \And
 Kangjie Yang \\
  \texttt{yangkj@motern.com} \\
  \And
 Zongyang Liu \\
  \texttt{lzy@motern.com} \\
  \And
 Xin Mao \\
  \texttt{maoxin@motern.com} \\
  \\
  Shenzhen Motern Technology Co., Ltd.
}

\begin{document}
\maketitle
\begin{abstract}
This technical report presents the implementation of a state-of-the-art video encoder for video-text modal alignment and a video conversation framework called HiLight, which features dual visual towers. The work is divided into two main parts: 1.alignment of video and text modalities; 2.convenient and efficient way to interact with users. Our goal is to address the task of video comprehension in the context of billiards. The report includes a discussion of the concepts and the final solution developed during the task's implementation.
\end{abstract}


\section{Introduction}
A recent line of MLLMs work\cite{Video-LLaVA,videogpt+} has started to explore introducing video modality. On the one hand, video contains more information than images because of the time-dimensional modelling capability, and video is much better in capturing events. On the other hand video cameras are everywhere around us as a common and low-cost electronic surveillance tool. However, how to transform advanced academic results or models into beneficial and practical products is a subject of much greater concern to industry. According to business requirement of the company, our team will focus on a challenging task, video understanding of billiards indoor scenes. 

We propose HiLight, a video chat model. HiLight is constructed by the two stages, each of which is required to accomplish respective target performance:

\begin{enumerate}
\item Video and Text Modality Alignment Stage: This stage requires aligning the video modality with the text modality as well as a strong ability to detecting small objects.
\item Visual-Language Model Fine-Tuning Stage: This stage allows users to interact in conversation with open vocabulary and requires the model to be able to make logical reasoning to some degree.
\end{enumerate}

\section{First Stage - Video Encoder}
\subsection{Improvements to the Video Encoder}
Most VLMs utilize CLIP model\cite{CLIP}, which benefit from the massive scale of noisy web image-text data. However, it lack explicit temporal context, which can be important in videos with intricate action sequences. To align the video and text modalities, we select Microsoft's CLIP-ViP\cite{CLIP-ViP} as our video encoder. This is an excellent work on contrastive learning pre-training models. By introducing proxy mechanism, it effectively preserves the capabilities of the image-based CLIP\cite{CLIP} pre-training model while also capturing the temporal continuity between video frames.

To the understanding the video modality in the context of billiards, it is necessary to be precise about tracking each ball's trajectory and interactions. Therefore, we would like to introduce explicit spatial relationship modeling, mapping the object features in the original input directly to corresponding spatial positions in the Vision Transformer (ViT)\cite{ViT} patches. How can we exert this explicit modeling constraint while aligning the video and text modalities? We found that Google DeepMind’s SPARC\cite{SPARC} can effectively provide a solution. They proposed to add a local loss between each patch from the Vision Encoder and each token from the Text Encoder, SPARC ensures that every token has at least one semantically similar patch.

\subsection{Experiments on the Video Encoder}
However, However, there are many details to be considered when adding a local loss to the already pre-trained CLIP-ViP smodel\cite{CLIP-ViP}, such as whether it should be aligned at the same layer of output as the original global loss?

If not aligned at the same layer, local loss is applied in the light layer between the embedding and the encoder, while global loss is applied after the original encoder, there are two considerations:

\begin{enumerate}
\item Embedding Layer Constrains and Encoder Block Constrains: Consider that local loss constrains the embedding layer, and global loss constrains the encoder block after the embedding. This situation is similar to gradually freezing light layer weights while training downstream tasks, which could accelerate convergence.
\item Constraint Capability of the Embedding Layer: Is the constrainability of the embedding layer sufficient for the level of local loss?  It is likely that the embedding layer is not deep enough for the network to be able to fit such a fine-grained and very high level of local loss  (this can be verified by the decline of the local loss in actual training).
\end{enumerate}

If aligned at the same layer, with both local loss and global loss applied in the deep layers after the encoder, there are three considerations:

\begin{enumerate}
\item Mutual Constraint and Consistent Gradient Trends: Both losses constrain each other and act on the same features, leading to consistent gradient trends and avoiding conflicting learning directions across different layers.
\item Implicit Fine-Grained Learning in Deep Layers: Deep layers are not suitable for fine-grained learning, as patches and tokens lack clear semantic distinctions. Fine-grained learning at this level is implicit rather than explicit. This raises a direct question: Is explicit textual masking (language mask) still necessary for fine-grained contrastive learning of implicit features?
\item Effect of Sparsity in Deep Layers: Is the effect of sparsity in learning fine-grained local loss in deep implicit modeling consistent with that in light explicit modeling? It is obviously not consistent, so is sparsification still necessary in deep local loss? If so, how to pick the sparsification threshold and is it still 1/patch?
\end{enumerate}

Therefore, we designed four sets of experiments to verify the effects of different levels of local loss and global loss:

\begin{enumerate}
\item Local loss calculated using visual features and text features before the projector layer, without language mask.
\item Local loss calculated using visual features and text features before the projector layer, with language mask.
\item Local loss calculated using outputs of CLIPTextEmbeddings and CLIPVisionViPEmbeddings, with language mask.
\item In this setting, we would like to solve two problems at once: insufficient learning in light layer computation of loss, and deep layer computation of loss can't constrain the semantic boundary between patches and tokens. The core idea is to retain and pass the semantic boundaries of text/image through each layer of the encoder. Specifically, Local loss is calculated for each hidden state layer of the CLIP Encoder, with increasing weight from 0.1 to 1 through the layers, preserving local loss completely at the final layer.
\end{enumerate}

In our experiments, we observed that, under the same training cost, the method of calculating local loss after the encoder but before the projector layer demonstrated clear spatial correspondence in visualizations (since our visualize feature maps are GIFs, it is inconvenient to display here). Additionally, as to whether a language mask is required or not, we concluded from our experiments:

We experimented on some specific samples, and it is a noticeable fact that the setting with language mask performs much better than the setting without one.

A sensible explanation is that since there is no language mask, abstract terms such as "start of the ball game" require each token to correspond to each patch when calculating the loss between tokens and patches. But  such abstract tokens do not seem to have any patches that can correspond to them well, especially when patches are required to capture more spatial semantic information during local loss training. As training progresses, the abstract semantics are worn out. However, In the case of language mask, masked tokens can act as global tokens, preserving information as they do not contribute to the loss - retaining the abstract information of the tokens involved in the computation.

This explanation is supported by visualizations of the vision encoder, where adding a language mask in the Text Encoder significantly impacts the attention maps of the vision encoder. Without the language mask, the loss computation forces tokens with abstract semantics to align with patches, causing abstract information in tokens to be lost and patches in the vision encoder to acquire pseudo-global information to align with tokens. Consequently, more meaningless noise is observed in visualizations of the control group without language masks, attributed to forced alignment of abstract tokens with patches, creating pseudo-global patches.

\begin{figure}
  \centering
  \includegraphics[width=0.65\textwidth]{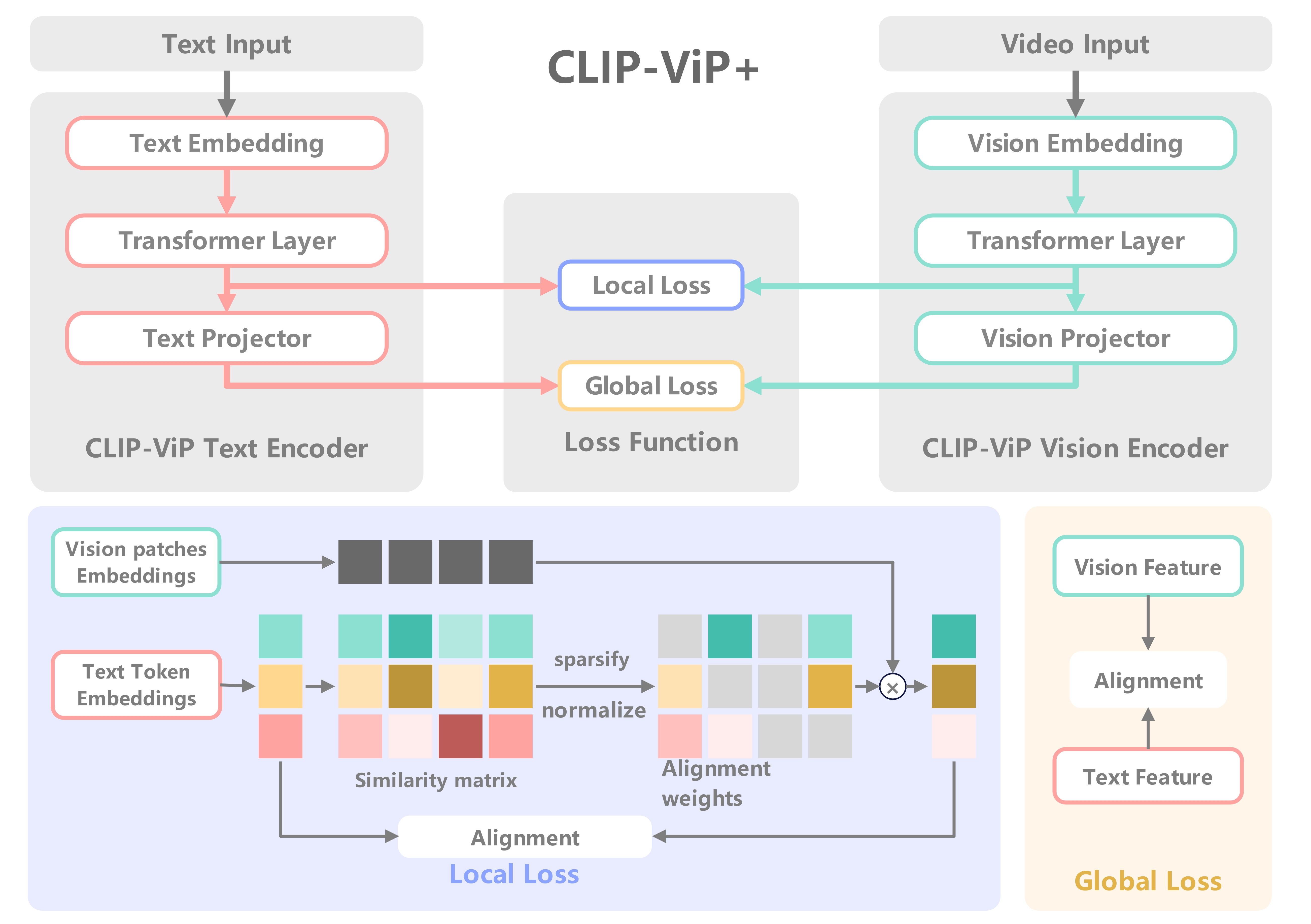}
  \caption{A improved CLIP-ViP structure -- CLIP-ViP+,  where global loss is from the contrastive learning of CLIP-ViP and introduced local loss is from the SPARC method\cite{SPARC}.}
  \label{fig:fig1}
\end{figure}

Therefore, the second experimental setting will be used to train the fine-grained aligned vision encoder, which we call CLIP-VIP+, the gray part of Figure 1\ref{fig:fig1}.It is improved from CLIP-VIP\cite{CLIP-ViP}.

\section{Second Stage - Visual Language Model}
\subsection{Background on VLM}
Based on a well-trained video encoder, we proceed to construct the framework for a Visual Language Model (VLM). At this stage, we draw insights from methodologies employed by many LLaVa-derived models\cite{LLaVA}. A common VLM implementation paradigm that allows the language model to do auto-regressive generation in the form of a vision tokens, where a robust vision encoder is significant. Therefore, more and more frameworks like Ferret-v2\cite{Ferret-v2}, Mini-Gemini\cite{Mini-Gemini} and Cambrian-1\cite{Cambrian-1}, etc. use multiple vision encoders to construct enhanced visual information input.

For instance, in Mini-Gemini\cite{Mini-Gemini}, the process of feeding visual information into the Language and Vision Model (LLM) involves two streams: a high-resolution feature extractor (HR) and a low-resolution feature extractor (LR). These extractors sample original and downsampled images, respectively, to obtain features with varying levels of information content. These features are then combined through cross attention, referred as "token mining" .

\subsection{Discussion on Dual-Tower Vision Encoder}
A concern is how to combine multiple vision extractors and explore what they might effect. 
In Mini-Gemini\cite{Mini-Gemini}, the authors consider the dual-tower structure to be able to merge the advantages of the high-resolution vision Encoder and the excellent text-aligned CLIP Vision Encoder. In Video-LLaVA\cite{Video-LLaVA}, the authors also used the image encoder as an auxiliary in the video processing. Cambrian-1\cite{Cambrian-1} explore the combinations multiple vision encoders, focusing on the advantages of integrating self-supervised models of the CLIP type for text sensitivity and SSL models of the DINOv2 type for observation of image details.

How to make the model assemble combine the advantages of multiple vision encoders rather than the disadvantages of these components is still a question to be explored. Cambrian-1\cite{Cambrian-1} presents a new design in this problem that promises to provide a powerful solution - the Spatial Vision Aggregator(SVA). However, it is certain that if there is a vision encoder that is equal to multiple virtues at once, there is no more necessity for complicated multi- tower structures and connector designs. But in the design of HiLight framework in this project, we don't afford to construct a versatile vision encoder with multiple advantages in the short term, so we still employ a dual-tower structure to enhance the feature extraction.

The two towers of our HiLight architecture are CLIP-ViP+ which are trained already before, and Long-CLIP\cite{Long-CLIP}, which provides long contextual modal alignment capability beyond CLIP. We hope to balance the excellent and complete spatial temporal consistency modelling of CLIP-ViP+ with the superior modal alignment capability of Long-CLIP over CLIP\cite{CLIP}. We would like to take one of video encoders is the fine-tuned CLIP-ViP+ with a finer-grained Local Loss ,and the other side of the dual-towers is a model comparable to or even better than CLIP to compensate for the loss of general capability in fine-tuning CLIP-ViP+.The Long-CLIP is a good choice.

\subsection{ HiLight Video Conversation Model}

\begin{figure}
  \centering
  \includegraphics[width=1\textwidth]{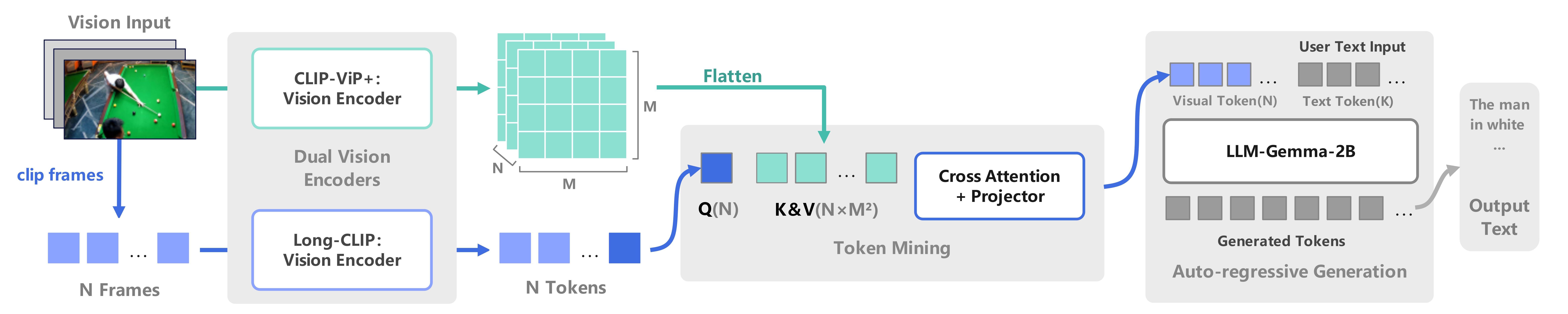}
  \caption{HiLight Dual-Tower VLM Framework. CLIP-ViP receives complete video features, while Long-CLIP receives fixed-sampled video frames. The features of each keyframe outputted by Long-CLIP serve as the query for Cross Attention with the complete video features from CLIP-ViP. Following a projection layer output vision tokens which are fed into the language model concat with the user's text input. }
  \label{fig:fig2}
\end{figure}

As shown in Figure\ref{fig:fig2}, our HiLight structure consists of a video input into two vision encoder inputs respectively. Where CLIP-ViP+ is for extracting the complete video continuity modelling and Long-CLIP is for modelling the keyframes  (which be sampled with constant frames or dynamically sampled). Then, the outputs of the two towers are fused with token mining to get the vision Token, and finally the vision token will be fed into the Gemma-2B\cite{Gemma} language model, and the whole process is completed.

\subsection{Training of VLM}
Our training steps are consistent with those of most such architectures and consist of two stages:

\begin{enumerate}
\item Modality Alignment Stage: In the first stage of VLM training, we utilize publicly available video Q\&A datasets to adequate train our token mining component. This stage aims to directly implement  the bridge from the visual features of the two towers to into vision token as input of the LLM.

\item Instruction Tuning Stage: The second stage of VLM training involves fine-tuning LLM using a custom dataset from billiards scenes. This stage is customised to perfect the model for specific scenario application.
\end{enumerate}

\begin{figure}
  \centering
  \includegraphics[width=0.65\textwidth]{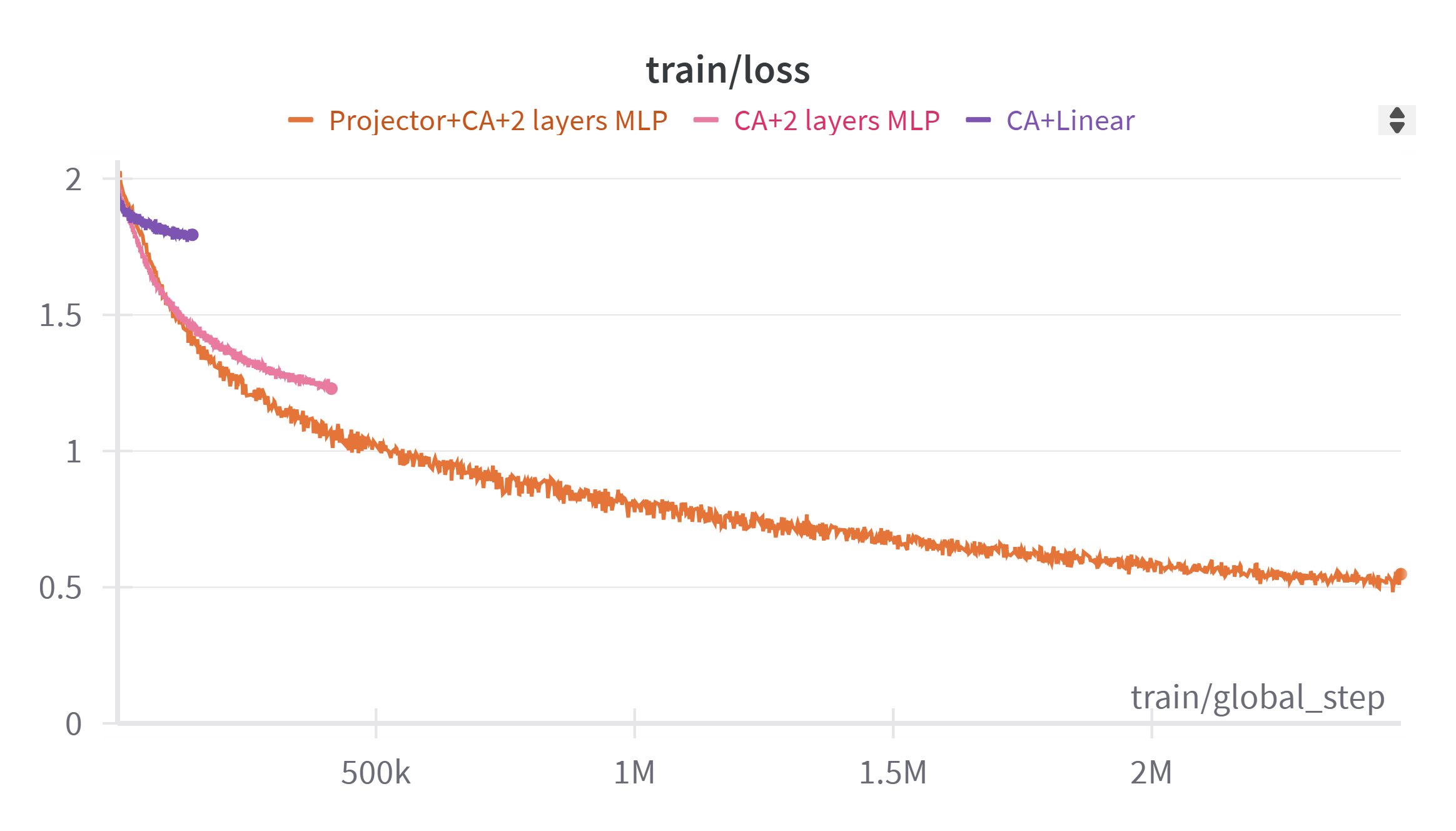}
  \caption{the first stage of VLM training, token mining training loss.}
  \label{fig:fig3}
\end{figure}

In the first stage, we conducted two structures for training token mining:

the first structure: one layer of Cross Attention  + one Linear layer, training 70 epochs purple in Figure\ref{fig:fig3};

the second structure: one layer of Cross Attention + two Linear layers + ReLU activation, trained for 200 epochs pink in Figure\ref{fig:fig3}.

It is obvious in the figure that the more layers of token mining, the lower loss value. The second structure, with an additional linear layer and two layers of ReLU activate function compared to the first set, bring down the loss from 1.7 to 1.3.

Next, we introduced the third training setting: the outputs from the dual towers each pass through one layer of Projector + one layer of Cross Attention + two layers of Linear + Relu activate function. The result is shown in orange in Figure\ref{fig:fig3}.

\section{Conclusion}
In this work, we implement an improved architecture CLIP-ViP+ and applied it in the HiLight video conversation framework. Overall, although this piece of work is essentially a combination of multiple existing jobs, it also requires some grounding: exploring the specificity of the subject, knowledge of multiple existing technology, the setting up of an experimental proposal and the determination of the feasibility of tasks.

Through this work, we got a complete exposure to every aspect of the visual language model, and accumulated valuable technical experience for the team from scratch. Following up on the HiLight model, we also hope to further improve the relevant details in the architecture:

\begin{enumerate}
\item Training a vision encoder that can satisfy both good modal alignment, high-resolution detail sensitivity, and long context support with low computational, and replacing the complicated dual tower structure with it.
\item Given video modality takes up more vision tokens relative to image modality, there is an urgent necessity to employ language models with low-cost computation to suit long text computation for long videos.
\item 
A better engineering support for multiple video inputs in a single round of conversation and the understanding of multiple videos.
\end{enumerate}

\section*{Author Contributions}

\textbf{Zhiting Wang} was responsible for algorithm development and made equal contributions. \textbf{Qiangong Zhou} participated in algorithm development and made equal contributions. \textbf{Kangjie Yang} provided data support. \textbf{Zongyang Liu} supervised the project. \textbf{Xin Mao} supervised the project.

\bibliographystyle{unsrt}  
\bibliography{template}  






\end{document}